\documentclass[runningheads]{llncs}

\usepackage{amssymb}
\usepackage{graphicx}
\usepackage[cmex10]{amsmath}
\usepackage{lineno}
\usepackage{array}

\usepackage{url}
\urldef{\mailsa}\path|{sxd7257, calbme}@rit.edu|

\def\registered{{\ooalign{\hfil\raise .00ex\hbox{\scriptsize R}\hfil\crcr\mathhexbox20D}}}

\begin{document}

\mainmatter  


\title{Integrating Atlas and Graph Cut Methods for LV Segmentation from Cardiac Cine MRI}

\titlerunning{\it LV Segmentation from Cardiac Cine MRI}

%
%
%
\author{Shusil Dangi${^1}$ \and Nathan Cahill${^{1,3}}$ \and Cristian A. Linte${^{*1,2}}$}
\authorrunning{Dangi, S. {\it et al.}}
%
\institute{${^1}$Chester F. Carlson Center for Imaging Science, ${^2}$Biomedical Engineering, \\
${^3}$Center for Applied and Computational Mathematics,\\ Rochester Institute of Technology, Rochester NY USA \\
}
%
\maketitle
%
\begin{abstract}
Magnetic Resonance Imaging (MRI) has evolved as a clinical standard-of-care imaging modality for cardiac morphology, function assessment, and guidance of cardiac interventions. All these applications rely on accurate extraction of the myocardial tissue and blood pool from the imaging data.
Here we propose a framework for left ventricle (LV) segmentation from cardiac cine-MRI. First, we segment the LV blood pool using iterative graph cuts, and subsequently use this information to segment the myocardium. We formulate the segmentation procedure as an energy minimization problem in a graph subject to the shape prior obtained by label propagation from an average atlas using affine registration.
The proposed framework has been validated on 30 patient cardiac cine-MRI datasets available through the STACOM LV segmentation challenge and yielded fast, robust, and accurate segmentation results.
\end{abstract}

\section{Introduction}
\label{sec:Intro}
The World Health Organization (WHO)\footnote{\url{http://http://www.who.int/mediacentre/factsheets/fs317/en/}} estimated 17.5 million deaths from cardiovascular diseases in 2012, representing $31\%$ of all mortalities, rendering cardiovascular conditions the main cause of death globally. Hence, the timely diagnosis and treatment follow-up of these pathologies is crucial. High image quality, good tissue contrast, and no ionizing radiation has established MRI as a standard clinical modality for non-invasive assessment of cardiac performance. Cardiac contractile function quantified via the systolic and diastolic volumes, ejection fraction, and myocardial mass represents a reliable diagnostic value and can be computed by segmenting the left (LV) and right (RV) ventricles from cardiac cine-MRI. Although manual delineation of the ventricle is deemed as the gold-standard approach, it requires significant time and effort and is highly susceptible to inter- and intra-observer variability. These limitations suggest a need for fast, robust, and accurate semi- or fully-automatic segmentation algorithms.

Various segmentation techniques for cardiac MR images have been proposed in the literature \cite{Petitjean2011}. The image-based approaches with weak or no prior information, such as thresholding, edge-based and region-based approaches, or pixel-based classifications methods, require user interaction for proper segmentation of the ill-defined regions. On the other hand, shape prior deformation models, active shape and appearance models, and atlas-based approaches are more likely to overcome this problem
at the expense of manually building a training set.

Multi-atlas based approaches have shown promising results in biomedical image segmentation \cite{Iglesias2015}. However, they rely on a number of computationally demanding and time limiting nonrigid image registration steps followed by label fusion. Hence despite its accuracy, it has experienced minimal to no adoption in actual clinical applications primarily due to its complexity, high dependence on parameters variability, and computational demands. 

On the other hand, combinatorial optimization based graph-cut techniques are fast and guaranteed to produce results within a known factor of the global minimum, for some special classes of functions (termed as regular functions) \cite{Kolmogorov2004} and have proved to be powerful tools for image segmentation. Moreover, adding a shape constraint into the graph cut framework has been shown to improve the cardiac image segmentation results significantly \cite{Grosgeorge2013,Freedman2005,Mahapatra2013}. However, these methods require a manual input to introduce a shape constraint at the right location in the image.



In this work, we leverage the performance of the graph cut framework and augment it by incorporating shape constraints in the form of an average atlas-based segmentation of the anatomy whose label was generated and propagated using a single affine registration. Subsequently, we iteratively refine the segmentation using techniques similar to those described in \cite{Slabaugh2005,Vu2008}, to obtain an accurate and robust segmentation of the myocardium. Hence, we do not require any manual input to introduce shape constraint into the graph-cut framework and simultaneously take advantage of the prior knowledge in the form of atlas-based segmentation requiring affine as opposed to nonrigid registration, which is more computationally efficient and less sensitive to parameters variability.

%

\section{Methodology}
\label{sec:Methods}

Whole heart cine-MRI images are generated by stacking 2D+T short-axis slices acquired during a single breath hold. Since this acquisition approach introduces an intensity difference between the slices, as well as slice misalignments, we can follow one of two approaches to segment tha data: one approach is to implement a slice motion correction protocol to realign the slices into a coherent 3D volume. The other approach, also implemented here, resorts to slice-wise processing and segmentation instead of a 3D segmentation.

Another challenge is the ill-defined contrast of the LV myocardium in MR images, which makes the image-driven segmentation difficult. As such, to obtain better segmentation of the apical and basal regions,
we exploit the prior knowledge in the form of an average atlas. The proposed methodology formulates the segmentation problem in the context of a graph based energy minimization framework. The blood pool is first segmented using an iterative graph cut technique; then, this information is used to segment the myocardium.

\subsection{Data Preprocessing}
\label{ssec:DataPreprocessing}
This study is conducted on 30 cardiac cine-MR images taken from the DETERMINE \cite{kadish2009} cohort available as a part of the STACOM Cardiac Atlas Segmentation Challenge Project database\footnote{\url{http://www.cardiacatlas.org}}. 
The semi-automatically segmented images obtained by applying the method described in \cite{Li2010} accompany the dataset and serves as gold-standard for assessing the proposed segmentation technique.

We select a reference patient volume with good contrast, average size, and preferred LV-RV orientation. All patient volumes are rotated about the z-axis (i.e., slice-encoding direction) to roughly align their orientation with that of the reference patient using the DICOM Image Orientation Patient (IPP) field. The region of interest (ROI) (in the xy-plane) enclosing the left and right ventricles is extracted using the method described in \cite{Kfir2016} by correlating the 2D motion images generated from the 3D volumes across the cardiac cycle. The only manual input required by our algorithm is the start and end slices of the LV, such that the ROI is restricted in the z-direction, preventing over/under segmentation of slices that do not belong to the desired anatomy. The patient volumes are cropped to the above ROI, and, to compensate for any intensity differences (due to the slice-wise acquisition), each slice is normalized (0-255) prior to further processing.

\begin{figure}[!t]
\centering
\includegraphics[width = \columnwidth]{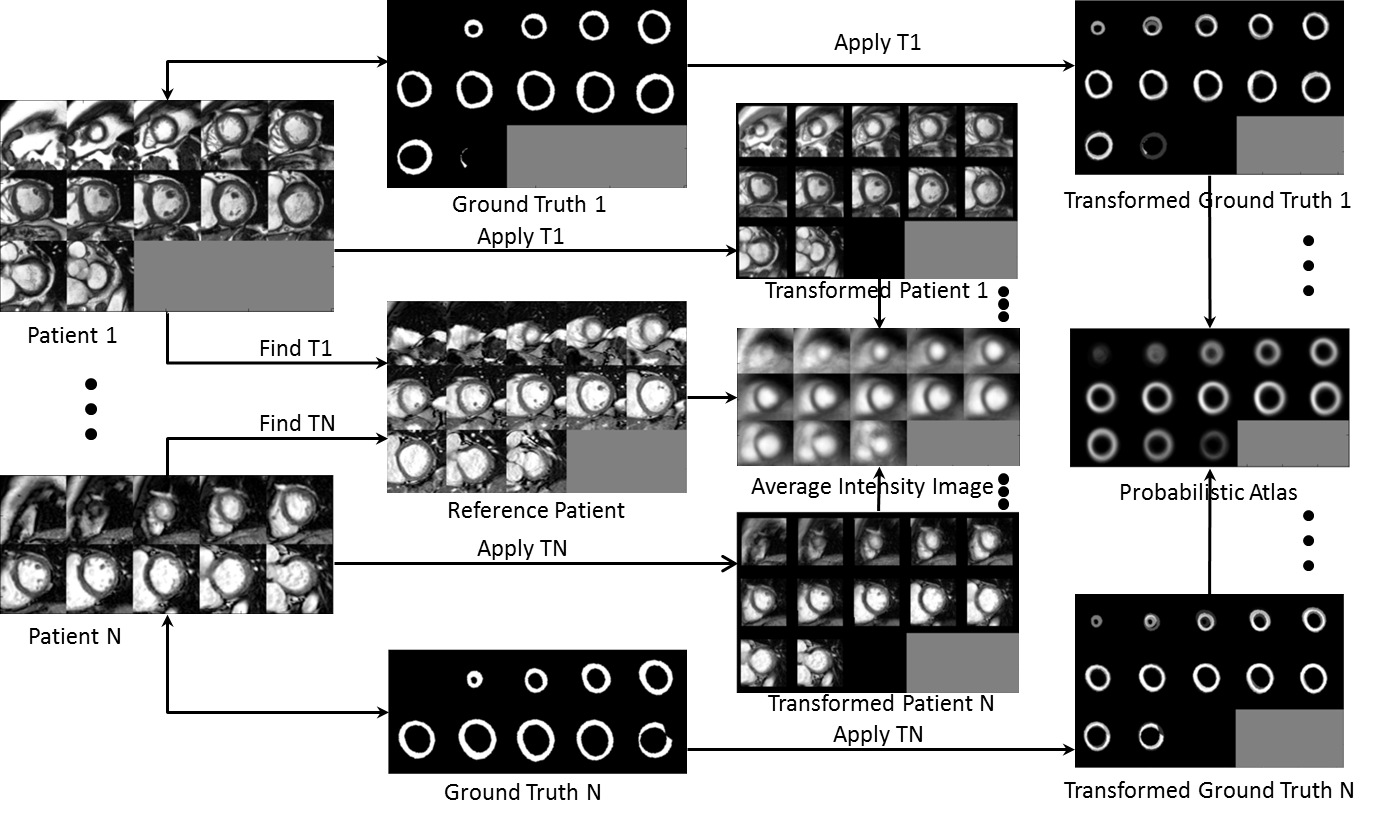}
\caption{\scriptsize{All patient images are affinely registered to the reference patient, and the obtained optimum transformation is applied to the corresponding ground truth images. An average intensity image is obtained by averaging the intensities of all transformed patient images, while, the averaging of the transformed ground truth images yields a probabilistic atlas.}}
\label{fig:AtlasGeneration}
\end{figure}

\subsection{Atlas Generation}
\label{ssec:AtlasGeneration}

The cropped 3D volumes (at the end diastole phase) for all patients are first histogram matched and then affinely registered to the reference patient image volume using the intensity based Nelder-Meade downhill simplex algorithm \cite{Nelder1965} available in SimpleITK. The resulting 3D affine transforms are applied to the respective ground truth segmentations. The transformed volumes and transformed ground truths are then averaged to obtain an average appearance atlas and a probabilistic atlas, respectively ({\bf Fig.~\ref{fig:AtlasGeneration}}). 

The average appearance atlas is registered to a test volume using intensity-based affine registration. The resulting registration transformation is used to transform the myocardial probabilistic label to the test data, which, in turn, serves as a shape constraint for the graph cut framework.

\subsection{LV Blood Pool Segmentation using Iterative Graph Cuts}
\label{ssec:BPSeg}
To leverage the 3D LV geometry, we use the blood pool (BP) segmentation of a given slice to help refine the BP ROI in the neighboring slices. As such, we first segment the BP from the mid-slice, followed by its neighboring slices, and proceed accordingly, until the complete volume is segmented.

\subsubsection{Intensity Distribution Model:}
\label{sssec:IDM}
The myocardium probability map for each slice is normalized and inverted to produce the probability map corresponding to the blood pool (BP) and background (BG). The resulting BP/BG probability map is thresholded at 0.5 and the inner connected component is isolated to obtain the high confidence BP ROI. Otsu thresholding \cite{Otsu1979} is applied within this ROI to obtain the initial BP region. The intensity values within this extracted BP region are then fitted to a Gaussian distribution to generate the BP intensity model. 

\begin{figure}[!h]
	\centering
	\includegraphics[width = \columnwidth]{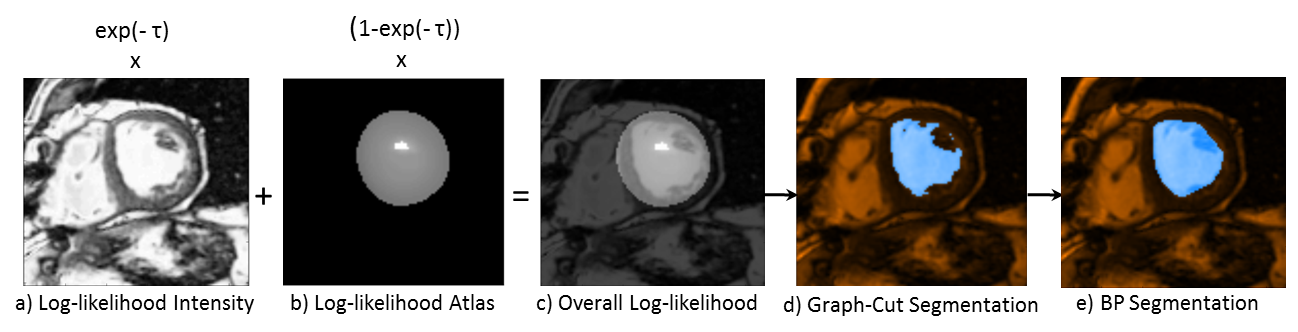}
	\caption{\scriptsize{Log-likelihood image obtained from: a) the intensity distribution model, b) the BP probabilistic map, c) weighted sum of (a) and (b); d) segmentation obtained from graph cut, e) convex hull of (d) yields the BP segmentation.}}
	\label{fig:BPsegment}
\end{figure}

A binary mask enclosing the myocardium is obtained by thresholding the myocardial probability map at a very small value (i.e. 0.1). Holes in the binary mask are filled to obtain a ROI enclosing the BP, myocardium, and BG. To generate the BG intensity model, 
we fit the intensity values within the ROI, excluding the initial BP region, to a Gaussian Mixture Model (GMM) comprising two Gaussians. {\bf Fig. \ref{fig:BPsegment}a} shows the resulting BP log-likelihood map.

Note that we propose the Gaussian distribution for modeling intensity noise in MR images instead of a more appropriate Rician distribution \cite{Gudbjartsson1995}; this simplifies our model and is a good approximation when the signal-to-noise ratio is high.

\subsubsection{Blood Pool/Background Probabilistic Map:}
\label{sssec:BP/BGMap}
To obtain a ROI that includes the myocardium and BP, we threshold the myocardial probability map at 0.5, fill in the blood pool, and erode the resulting ROI by $15\%$ (selected empirically) of the radius of its smallest circumscribed circle to obtain the BP ROI. The BP/BG probability map masked by the BP-ROI represents the BP probability map, and its inverse represents the BG probability map. {\bf Fig.~\ref{fig:BPsegment}b} shows the BP log-likelihood map.

\subsubsection{Graph-Cut Segmentation:}
\label{ssec:GCSeg}
We construct a graph with each node (i.e., pixel) connected to its east, west, north, and south neighbors. Two special terminal nodes representing two classes --- the source (blood pool), and the sink (background) --- are added to the graph and all other nodes are connected to each terminal node. The segmentation is formulated as an energy minimization problem over the space of optimal labelings $f$:
\begin{equation}
\label{eq:1}
E(f)~=~\sum\limits_{p \in \mathcal{P}} D_p(f_p)~+~\sum\limits_{\{p,q\} \in \mathcal{N}} V_{p,q}(f_p,f_q),
\end{equation}
where the first term represents the data energy that reduces the disagreement between the labeling $f_p$ given the observed data at every pixel $p \in P$, and the second term represents the smoothness energy that forces pixels $p$ and $q$ defined by a set of interacting pair $\mathcal{N}$ (in our case, the neighboring pixels) towards the same label. 

The data energy term is represented by the terminal link (t-link) between each node and the source (or sink), which is defined as the weighted sum of the log probabilities of the intensity distribution model and the probabilistic map corresponding to the BP (or BG):
\begin{multline}
\label{eq:2}
D_p(f_p)=exp\left(-\tau \right)*[-lnPr(I_p|f_p)]+(1-exp\left(-\tau \right))*[-lnPr(f_p)]
\end{multline}
where, $\tau$ is the iteration number, $Pr(I_p|f_p)$ is the likelihood of observing the intensity $I_p$ given that pixel $p$ belongs to class $f_p$, and $Pr(f_p)$ is the prior probability for class $f_p$ obtained from the BP/BG probability map. The log-likelihood difference between BP and BG labels for $\tau=1$ is shown in {\bf Fig.~\ref{fig:BPsegment}c}. The intensity likelihood term (first term) allows the expansion of the BP region in the first few iterations, whereas the prior probability map (second term) restricts its ``spilling" (due to over-segmentation) in subsequent iterations.

The smoothness energy term is computed over the links between neighboring nodes (n-links), which are weighted based on their intensity similarity:
\begin{equation}
\label{eq:3}
V_{p,q}(f_p,f_q) = 
\begin{cases}
\tau*exp\left(-\frac{|I_p-I_q|}{\tau}\right) & \text{if}~f_p=f_q\\
0 & \text{if}~f_p \neq f_q
\end{cases}
\end{equation}
where $I$ is the pixel intensity. To avoid the ``spilling" of the BP into the myocardium or BG, the smoothness term changes with each iteration, such that, in order for the neighboring pixels to be assigned to the same label during the current iteration, their intensities must be closer than in the previous iteration.

Once weights are assigned to all edges in the graph, the minimum cut equivalent to the maximum flow is identified via the $\alpha$-expansion algorithm described in \cite{Boykov2001}. This approach yields the labeling (graph-cut) that minimizes the global energy of the graph that corresponds to the optimal segmentation ({\bf Fig.~\ref{fig:BPsegment}d}). 
Lastly, the convex hull applied to the graph-cut result constitutes the final BP segmentation, such that, the papillary muscles are included within the BP ({\bf Fig.~\ref{fig:BPsegment}e}).

\begin{figure}[!h]
	\centering
	\includegraphics[width = \columnwidth]{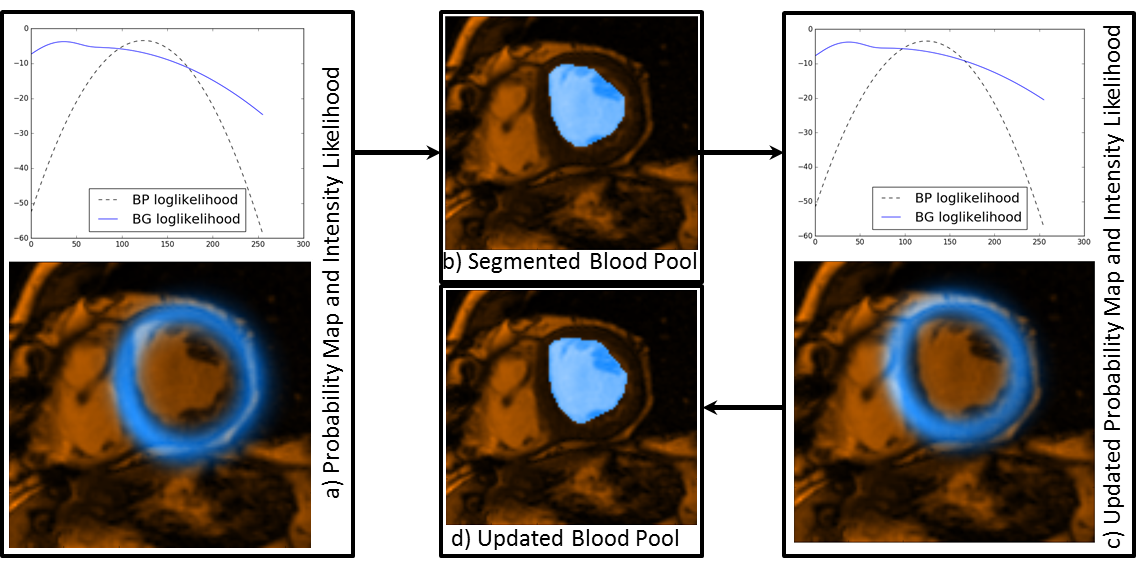}
	\caption{\scriptsize{a) Probability map and intensity distribution model for current iteration, b) BP segmentation obtained from graph cut using (a), c) updated probability map and intensity distribution model obtained using (b), d) new BP segmentation obtained from graph cut using (c).}}
	\label{fig:IterRefinement}
\end{figure}

\subsubsection{Myocardial Probability Map Refinement}
\label{sssec:MyoRefine}
The myocardial probability map is thresholded at 0.5, and the inner hollow circular region representing the BP is extracted. The signed distance map corresponding to the boundary of the extracted BP region is affinely registered to the signed distance map generated from the boundary of the graph-cut extracted BP ({\bf \ref{ssec:BPSeg}}) segmentation. The optimum affine transformation that minimizes the sum of squared differences between the two distance maps is applied to the myocardial probability map, such that, it fits the shape of the segmented BP.

\subsubsection{Iterative Refinement:}
\label{sssec:IterativeRef}
The latest BP segmentation obtained from the graph cut is used to update the intensity distribution model. The refined myocardial probability map is used to construct a new BP/BG probability map. The pixels within the latest BP segmentation are assigned very high likelihood (for belonging to the BP), and hence their labels do not change. An updated BP segmentation is obtained via another graph cut operating on the new graph energy configuration. This iterative process is repeated until the changes in the affine transform parameters for the myocardium probability map are below a predefined threshold; this iterative process usually converges within three iterations. Upon convergence, 
the convex hull defined by the latest segmentation result constitutes the final BP segmentation. 
{\bf Fig.~\ref{fig:IterRefinement}} illustrates the iterative refinement process.

\subsection{Myocardium Segmentation}
\label{ssec:MyoSeg}
The information from the BP segmentation along with the refined myocardial probability map is used to segment the myocardium.

\begin{figure}[!htb]
\centering
\includegraphics[width = \columnwidth]{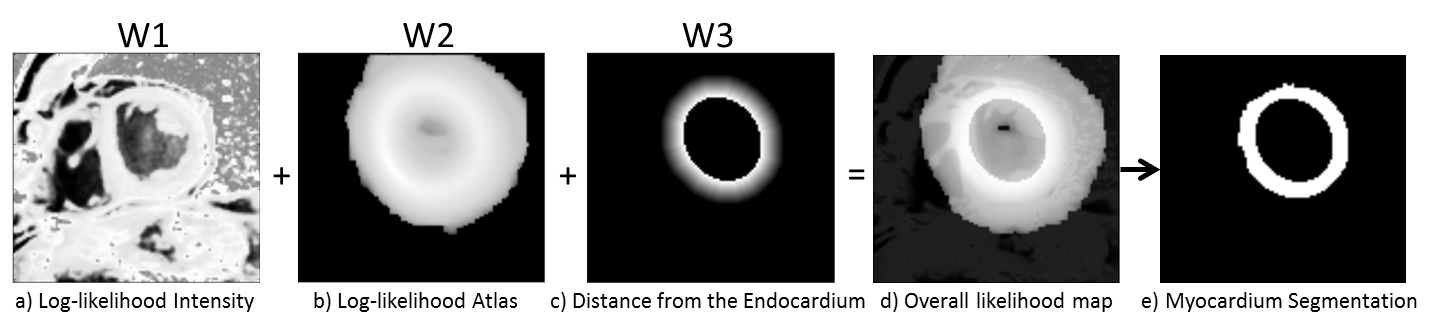}
\caption{\scriptsize{Log-likelihood image obtained from: a) the intensity distribution model, b) the refined myocardium probability map, c) distance from the endocardium; d) the weighted sum (w1, w2, and w3) of (a), (b), and (c), respectively; e) final myocardium segmentation obtained from graph cuts}}
\label{fig:MyoSegment}
\end{figure}

\subsubsection{Intensity Distribution Model:}
\label{sssec:IDM2}
We select a ROI in each slice based on the refined probability map, and we match the histogram of the pixel intensities within this ROI to the histogram of the mid-slice.
We select the mid-slices (i.e. no apical/basal slices) to obtain a single intensity distribution model for the whole volume. The intensities of the pixels within the refined myocardial mask with probability higher than 0.5 are fitted to a single Gaussian GMM to obtain the myocardium intensity distribution model. Similarly, the intensities of the remaining pixels are fitted to a three Gaussian GMM to obtain the BG intensity distribution model. {\bf Fig.~\ref{fig:MyoSegment}a} shows the log-likelihood map for the myocardium.


\subsubsection{Distance from the Endocardial Border:}
\label{sssec:distEndo}
The endocardial border is obtained from the outer edge of the final BP segmentation ({\bf Fig.~\ref{fig:BPsegment}e}). The knowledge that myocardium should be closer to the endocardial border is encoded in the data term represented by the truncated distance map (empirically selected as 10 pixels). This constraint increases the likelihood of pixels near the endocardial border to be labeled as myocardium, while reducing this likelihood for the pixels located further away. Furthermore, to prevent the BP region from being labeled as myocardium, it is assigned the lowest likelihood value ({\bf Fig.~\ref{fig:MyoSegment}c}).

\subsubsection{Graph-Cut Segmentation:}
A graph is constructed similar to the formulation described in {\bf \ref{ssec:GCSeg}}, but this time to classify the myocardial rather than blood pool pixels. The data term is defined as the weighted sum of the intensity distribution model, refined myocardial probability map (as described in {\bf \ref{sssec:MyoRefine}} and {\bf Fig.~\ref{fig:MyoSegment}b}), and the distance from endocardial border, with increasing relative influence, respectively. The smoothness term varies spatially according to the intensity difference between the neighboring pixes, as discussed in {\bf \ref{ssec:GCSeg}}. The minimum cut in the graph yields the final myocardium segmentation ({\bf Fig.~\ref{fig:MyoSegment}e}).

\section{Results}

The proposed algorithm was implemented in Python and required 45 seconds on average to segment the BP and myocardium from cine MRI volumes on an Intel$^\registered$ Xenon$^\registered$ 3.60 GHz 32GB RAM PC. 

\begin{figure}[!h]
\centering
\includegraphics[width = \columnwidth]{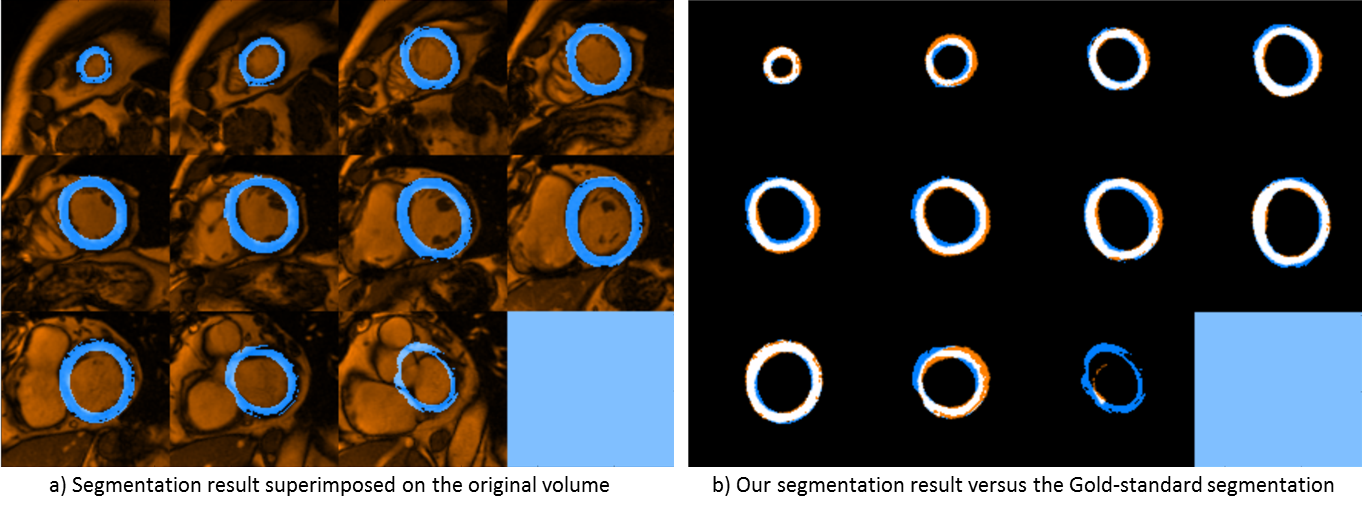}
\caption{\scriptsize{a) Final myocardium segmentation of all slices of a patient dataset (shown in blue) superimposed with the patient volume (shown in red); b) Final myocardium segmentation assessed against the provided gold-standard semi-automatic segmentation; white regions represent true positives, red regions represent false negatives, and blue regions represent false positives.}}
\label{fig:Results}
\end{figure}

Adhering to the collated results reported for the LV segmentation challenge in \cite{Suinesiaputra2014}, we evaluated our segmentation on 30 patient datasets according to the following metrics: dice index, jaccard index, sensitivity, specificity, positive predictive value (PPV), and negative predictive value (NPV) \cite{Suinesiaputra2012}. To maintain approximately equal number of myocardium and non-myocardium pixels for evaluation, such that the NPV conveys some useful information, we dilated each slice of the myocardium region, for the provided gold standard segmentation, by one fourth of the radius of the disk with equivalent area. The segmentation results for a patient dataset are overlaid onto each slice of the patient volume and shown in {\bf Fig.~\ref{fig:Results}a}. {\bf Fig.~\ref{fig:Results}b} shows a visual comparison of our segmentation results vis-\`{a}-vis the provided semi-automated segmentation serving as a gold-standard. The metrics are summarized in {\bf Table~1} for  all slices together, as well as for the mid-slices and apical/basal slices (first and last two slices, respectively) separately.

\scriptsize
\begin{table}
{\caption{Evaluation of our segmentation results against the provided gold-standard semi-automated segmentation for the mid-slices, apical/basal slices according to Dice Index, Jaccard Index, Sensitivity, Specificity, PPV, and NPV.}}
\label{tab:Table1}
\begin{center}
\begin{tabular}{|c||c|c|c|}
\hline
{\bf Assessment Metric} & {\bf Mid-Slices} & {\bf Apical/Basal-Slices} & {\bf All Slices} \\
\hline
\hline
{\bf Dice Index} & $0.811\pm0.068$ & $0.568\pm0.241$ & $0.740\pm0.180$ \\
\hline
{\bf Jaccard Index} & $0.687\pm0.091$ & $0.433\pm0.222$ & $0.613\pm0.183$ \\
\hline
{\bf Sensitivity} & $0.854\pm0.104$ & $0.596\pm0.268$ & $0.783\pm0.195$ \\
\hline
{\bf Specificity} & $0.788\pm0.103$ & $0.725\pm0.180$ & $0.770\pm0.134$ \\
\hline
{\bf PPV} & $0.789\pm0.079$ & $0.714\pm0.160$ & $0.767\pm0.114$ \\
\hline
{\bf NPV} & $0.866\pm0.086$ & $0.640\pm0.224$ & $0.800\pm0.174$ \\
\hline
\end{tabular}
\end{center}
\end{table}
\normalsize

\section{Discussion, Conclusion, and Future Work}
Our validation experiments show that the overall segmentation results are comparable to those reported in \cite{Suinesiaputra2012}. Specifically, the mean values for reported assessment metrics were: dice index --- 0.68 to 0.88, jaccard index --- 0.53 to 0.80, sensitivity --- 0.63 to 0.90, specificity --- 0.73 to 0.99, PPV --- 0.66 to 0.96, and NPV --- 0.81 to 0.94. However, it should be noted that the metrics reported in \cite{Suinesiaputra2012} were evaluated against the consensus segmentation estimated based on the participating seven raters (manual and automatic) obtained using the STAPLE algorithm on 18 test patient datasets, whereas our reported metrics were assessed against the provided semi-automated gold standard segmentation of 30 training patient datasets. Hence, the metrics provide only an approximate estimate of our algorithm's performance compared to the ones that participated in the challenge. Moreover, the average segmentation time of 45 secs per volume achieved using an unoptimized implementation in Python shows great potential for our algorithm to achieve, following minimal optimization, the near real-time performance demanded by some of the intended clinical applications.

Since the BP region in the mid-slices are better defined than in the apical/basal slices, the segmentation results are consistently better for the mid-slices. We also observed that the slice-wise processing and iterative refinement might compromise the segmentation of the apical/basal slices due to ill-defined BP regions, suggesting the need for special processing for these slices.

As part of our future work, we plan to automate the ROI detection in z-direction to eliminate the manual input required by our algorithm. In addition, instead of using a constant truncating endocardial distance constraint, we plan to use image-derived edge information to enable spatially varying truncating distances to improve the myocardium segmentation. Similarly, we will study the effect of selecting different thresholds for the probability maps, weight variability on the likelihood terms and, in turn, on the final myocardium segmentation. Lastly, we plan to extend the work and evaluate the segmentation performance on all 100 patient datasets and report performance according to the metrics outlined above.

\bibliographystyle{splncs}   
\bibliography{STACOM_2016}   

\end{document}